\documentclass{article}

\usepackage{arxiv}

\usepackage[utf8]{inputenc} 
\usepackage[T1]{fontenc}    
\usepackage{hyperref}       
\usepackage{url}            
\usepackage{booktabs}       
\usepackage{amsfonts}       
\usepackage{nicefrac}       
\usepackage{microtype}      
\usepackage{lipsum}		
\usepackage{graphicx}
\usepackage{natbib}
\usepackage{doi}

\title{MedDM:LLM-executable clinical guidance tree for clinical decision-making}
\author{ 
    \href{https://orcid.org/0000-0000-0000-0000}{\includegraphics[scale=0.06]{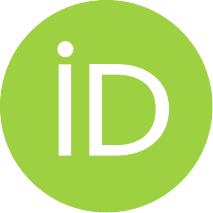}\hspace{1mm}Binbin Li} \\
	East China University of Science and Technology\\
	Shanghai, China \\
	\texttt{y80220053@mail.ecust.edu.cn} \\
	\And
	\href{https://orcid.org/0000-0000-0000-0000}{\includegraphics[scale=0.06]{orcid.pdf}\hspace{1mm}Tianxin Meng} \\
	East China University of Science and Technology\\
	Shanghai, China \\
	\texttt{y80220060@mail.ecust.edu.cn} \\
	\And
	\href{https://orcid.org/0000-0000-0000-0000}{\includegraphics[scale=0.06]{orcid.pdf}\hspace{1mm}Xiaoming Shi} \\
	Shanghai Artificical Intelligence Laboratory\\
	Shanghai, China \\
	\texttt{shixiaoming@pjlab.org.cn} \\
	\And
	\href{https://orcid.org/0000-0000-0000-0000}{\includegraphics[scale=0.06]{orcid.pdf}\hspace{1mm}Jie Zhai} \\
	East China University of Science and Technology\\
	Shanghai, China\\
	\texttt{zhbzj@ecust.edu.cn} \\
	\And
	\href{https://orcid.org/0000-0000-0000-0000}{\includegraphics[scale=0.06]{orcid.pdf}\hspace{1mm}Tong Ruan} \\
	East China University of Science and Technology\\
	Shanghai, China\\
	\texttt{ruantong@ecust.edu.cn} \\
}

\renewcommand{\shorttitle}{MedDM:LLM-executable clinical guidance tree for clinical decision-making}

\hypersetup{
pdftitle={MedDM:LLM-executable clinical guidance tree for diagnostic decision-making},
pdfsubject={Artificial Intelligence (cs.AI)},
pdfauthor={Li Binbin, Meng Tianxing，Shi Xiaoming},
pdfkeywords={LLM-executable clinical guidance tree, MedDM, differential diagnosis},
}

\begin{document}
\maketitle

\begin{abstract}
It is becoming increasingly emphasis on the importance of LLM participating in clinical diagnosis decision-making. However, the low specialization refers to that current medical LLMs can not provide specific medical advice, which are more like a medical Q\&A. And there is no suitable clinical guidance tree data set that can be used directly with LLM. 
To address this issue, we first propose LLM-executavle clinical guidance tree(CGT), which can be directly used by large language models, and construct medical diagnostic decision-making dataset (MedDM), from flowcharts in clinical practice guidelines. We propose an approach to screen flowcharts from medical literature, followed by their identification and conversion into standardized diagnostic decision trees. 
Constructed a knowledge base with 1202 decision trees, which came from 5000 medical literature and covered 12 hospital departments, including internal medicine, surgery, psychiatry, and over 500 diseases.Moreover,
we propose a method for reasoning on LLM-executable CGT and a Patient-LLM multi-turn dialogue framework.
\end{abstract}


\section{Introduction}
The \textbf{l}arge \textbf{l}anguage \textbf{m}odels (LLMs)~\citep{brown2020language} have shown great potential in various domains, including medical diagnosis~\citep{wang2023augmenting}, law~\citep{cui2023chatlaw}, education~\citep{joshi2023s}, thanks to their abilities to generate human-like responses by learning from tremendous online resources.
Among these application domains, LLMs for the medical domain gain increasing attention due to the wide range of practical needs and alluring technical potential to simplify the diagnostic process.

\begin{figure}
	\centering
        \label{fig:pdpc}
	\includegraphics[scale=0.18]{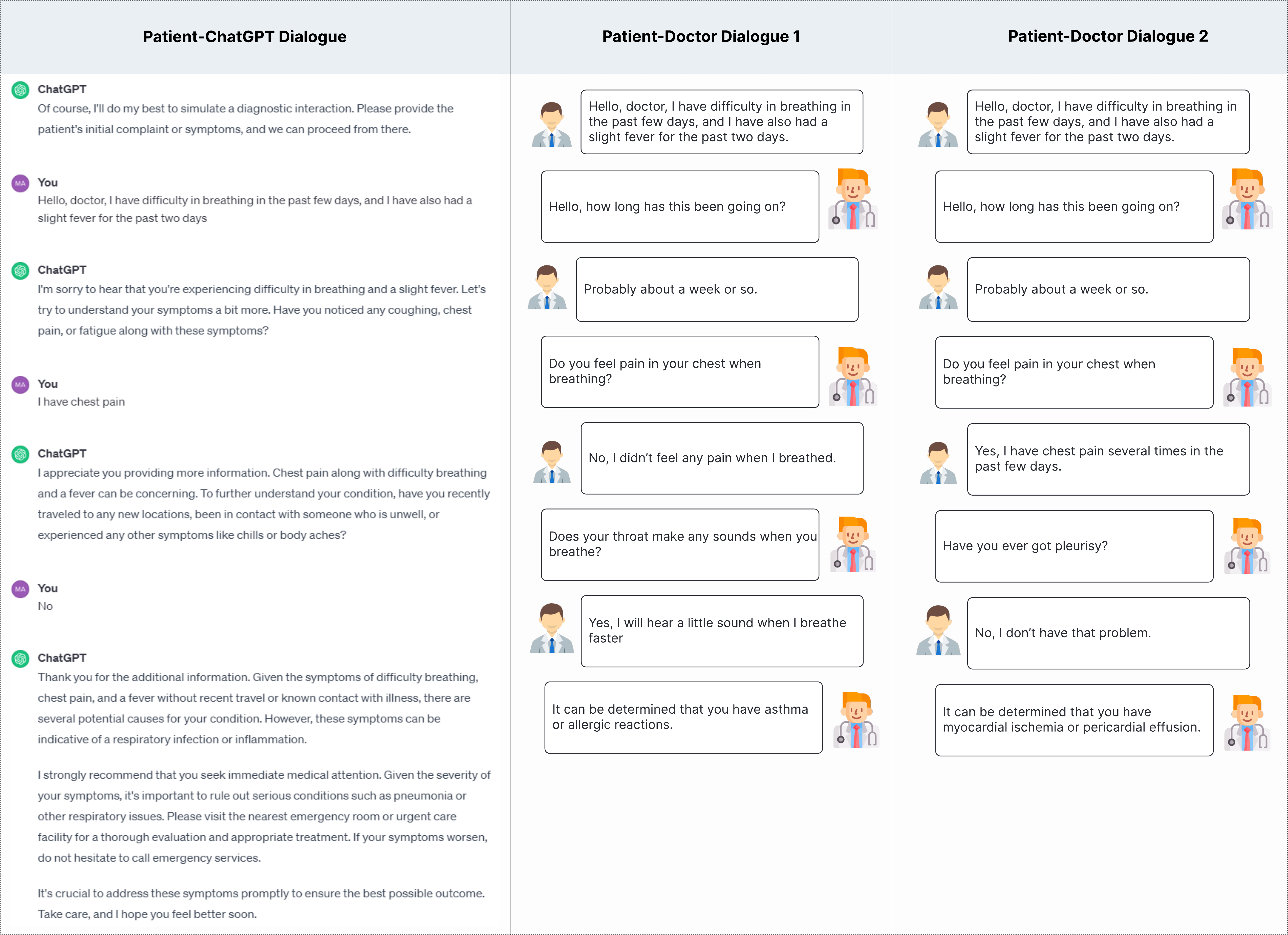}
	\caption{The difference between patients talking to doctors and ChatGPT}
\end{figure}

Despite the potential, medical LLMs suffer from low specialization for the medical diagnosis scenarios.
The low specialization refers to that current medical LLMs can not obtain complete information through multiple rounds of inquiries and then provide specific medical advice.
Instead, they are more like a medical Q\&A systems to just ask a few common symptoms and give a general response.
Examples from patient-doctor and patient-ChatGPT are shown in Figure~\ref{fig:pdpc}.
Compared with doctors, ChatGPT does not have enough differential diagnosis knowledge in terms of dyspnea, leading to low diagnosis quality.
Doctors, on the other hand, are able to further ask patients different symptoms depending on their health conditions, with their rich medical knowledge. 
Thus, the low specialization hinders practical application of medical LLMs in real medical scenarios.

To improve the medical specialization, diagnostic decision-making knowledge is necessary to conduct multi rounds of inquiries, and then make accurate diagnosis and provide personalized treatment. 
We propose an enhanced form of decisoion tree structure, namely LLM-executable clinical guidance tree(CGT), representing the nodes of the tree in natural language. 
It allows LLM to perform reasoning on CGT, thus generating more credible medical responses.
However, there is no publicly-available diagnostic decision-making dataset.
Constructing a diagnostic decision-making dataset is not trivial and challenging.
In this work, we first propose to construct the \textbf{me}dical \textbf{d}iagnostic \textbf{d}ecision-\textbf{m}aking dataset (MedDM), from flowcharts in clinical practice guidelines.
It is observed that there are a large number of flowcharts that precisely depict the decision-making process employed by doctors in disease diagnosis or treatment. 
These flowcharts inherently adhere to the structure of decision trees and can be seamlessly transformed into such representations.

Specifically, to conduct the dataset, 5,000 medical guidelines from various departments are collected. 
Then, layout analysis techniques are utilized to extract images from these medical books and a flowchart recognition model is proposed to pre-process the images. 
Subsequently, pictures which contain fewer than 8 basic flowchart graphs within the image are selected, and 2,300 images are obtained. 
Through manual verification, we further eliminate non-flowchart or medically unrelated images, ultimately yielding 1202 refined images.
To further identity the details in the flowchart, we developed a dataset and trained the Faster R-CNN~\citep{7485869} model to detect the fundamental structure of the flowchart, which achieves 98\% in accuracy. 
Then, to ascertain the starting and finishing points of the connecting lines, the contour detection~\citep{arbelaez2010contour} and DBSCAN~\citep{10.5555/3001460.3001507} clustering algorithm are utilized. 
Besides, to identify the text content in the flowchart, CnOCR~\citep{liu2021fake} tool are utilized. 
Finally, a checking team with 7 calibrators are established to cross-check and identify the completed flow chart to ensure that the relationship between nodes and connecting lines in the flow chart are correctly identified.


Furthermore, to generate informative and accurate responses for medical diagnosis dialog, we propose a decision-retrieval based generation framework to handle medical decision-making knowledge.

This work makes the following contributions:
\begin{itemize}
	\item We identify a new challenge, that is, in medical diagnosis scenarios, current medical LLMs lacks specialization.
 	\item To mitigate this challenge, we propose a novel decision tree structure, namely LLM-executable clinical guidance tree, and collect a new medical decision-making dataset (MedDM).
        \item We propose a decision-retrieval based generation framework to address the task. Experimental results show the effectiveness of the method.
\end{itemize}

\section{Related Work}
The clinical guidance tree \citep{turner2009abstraction} is an enhanced version of a decision tree employed in medical practice, utilizing a text-based representation method to define the guiding tree. \citep{li2022text2dt} recruited an annotation team to construct the Text2DT dataset based on rules authored by medical experts and extracted from medical literature, which employs a binary tree as its representation, with node information being represented by triplets.

Flowchart Recognition can be categorized into two major types: handwritten flowchart recognition and machine-generated flowchart recognition. Given that the medical literature collected for this study is in electronic PDF format, our focus is on machine-generated flowcharts. \citep{rusinol2012cvc} and \citep{morzinger2012visual} delineated flowchart recognition into two tasks: text recognition and graphic recognition. They trained two separate models for detection. In contrast, \citep{sun2022fr} proposed a multitask model based on the transformer architecture for identifying both text and graphics, achieving end-to-end flowchart recognition. However, the current effectiveness of existing flowchart recognition models in identifying connecting lines needs further improvement. Concurrently, we believe that mature text recognition tools such as CnOCR can assist in enhancing the accuracy of text recognition.

Clinical decision support system assists physicians in patient consultations, offering guidance and recommendations. The primary methods involve knowledge-driven and data-driven approaches. \citep{mei2011ocl}, based on clinical guidelines, developed a rule engine for managing chronic diseases. In order to provide personalized treatments, \citep{liu2017precision} grouped electronic medical record data of atrial fibrillation patients, demonstrating effectiveness in real cases. Moreover, \citep{zhao2020construction}, combining clinical guidelines and electronic medical record data, constructed a nested decision tree for grouping and treatment recommendations for patients with hyperthyroidism. Specifically, it involves firstly building a decision tree based on the clinical guides with rule-obeyed patient data, and then loading rule-uncovered patient data to further expand the tree.

\section{LLM-executable clinical guidance tree}

\subsection{Defining LLM-executable CGT}
Due to the increasing emphasis on the importance of LLM participating in clinical diagnosis decision-making, clinical guidance trees for direct use by LLM are becoming more and more urgently needed. Therefore, we propose a CGT representation that can be directly used by large language models----LLM-executable CGT. Contrasting with the traditional, regimented frameworks of clinical guidance and decision trees, we adopt natural language for a nuanced representation of node content. This strategy enhances clarity in conveying each node's information and offers direct compatibility for use as prompts in large language models.

\begin{figure}
	\centering
	\includegraphics[scale=0.26]{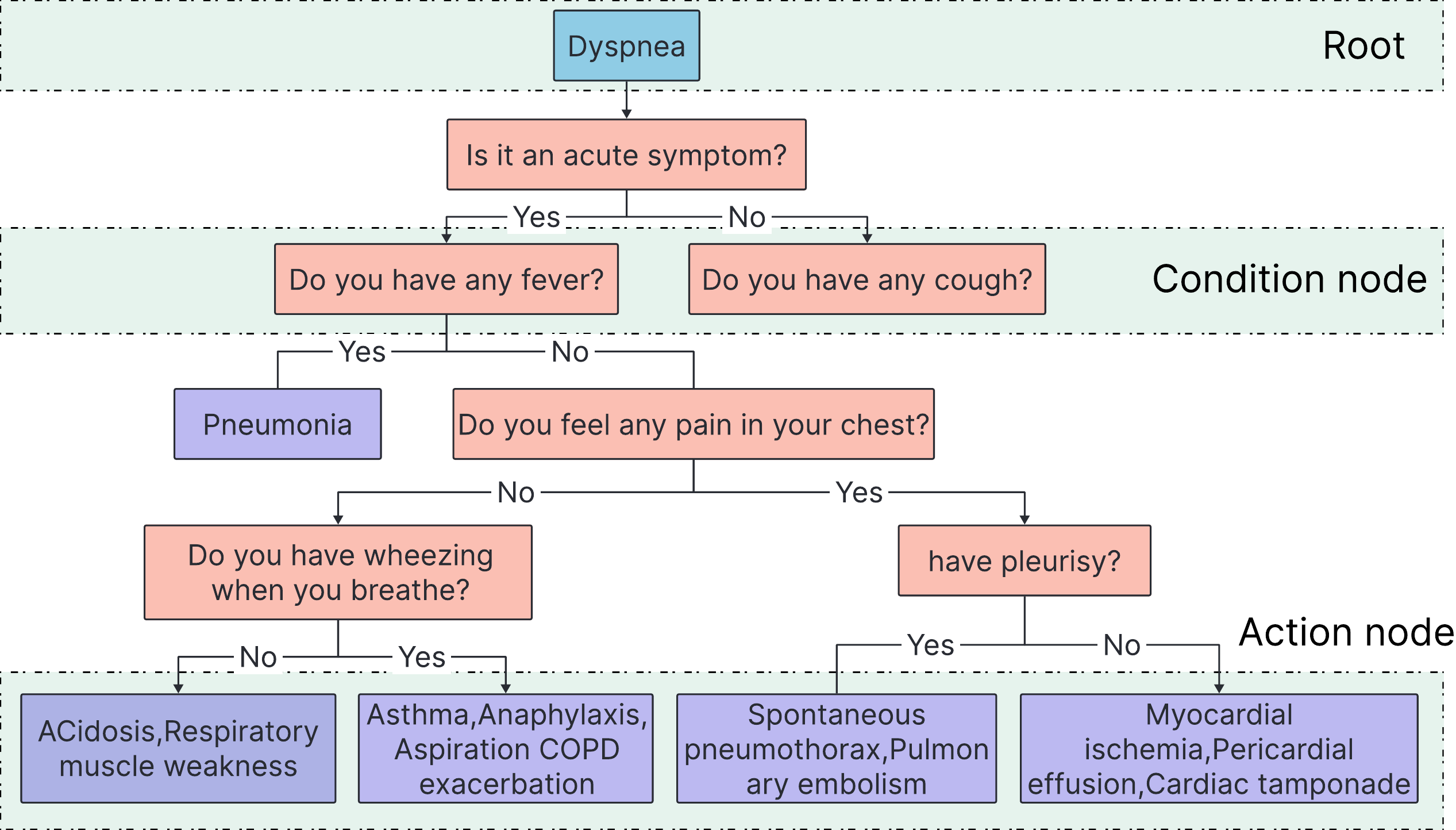}
	\caption{A differential diagnosis clinical  guidance tree with dyspnea symptoms as the chief complaint}
	\label{fig:fig3}
\end{figure}

We define the LLM-executable CGT enhanced, semi-structured representation of a decision tree for clinical decision making which is LLM-friendly and capable of being interpreted and executed. Using this representation, LLM can easily follow decision pathway to make medical decisions. Compared with the traditional CGT, our CGT is a multi-tree structure, which is only composed of root node, condition node and action node, as show in figure \ref{fig:fig3}. The root node represents the core symptoms or the name of the disease. Conditional nodes are the non-leaf nodes of the multi-branch tree, signifying the conditions that need to be evaluated during the decision-making process. The action nodes, as the leaf nodes, represent the outcomes of the final decisions.

\paragraph{Condition Node.}  Unlike the Text2DT \citet{zhu2022extracting}, which uses triples and logical relationships to represent node structures, our CGT nodes utilize natural language forms. This is because large language models today can comprehend content expressed in natural language and accurately perform simple reasoning tasks. Therefore, as show in figure \ref{fig:fig3}, in the "Dyspnea" clinical guidance tree, the question "Have any fever symptom?" is directly used as a condition.  This approach is not only intuitive for doctors and patients to understand, but it also enables the LLM model to perform reasoning on the CGT.

\paragraph{Action Node.}The leaf node, a pivotal component of the Clinical Guidance Tree (CGT), denotes the conclusion reached through CGT analysis. This conclusion manifests as either a definitive disease diagnosis or as prescribed examinations and treatment procedures necessary for the patient.

\subsection{CGT structure}
The tree is structured in a multi-branch format, enabling each decision node to support several child nodes. This configuration not only decreases the number of layers in the CGT but also consolidates scenarios sharing the same judgement criteria on one level. The result is a more streamlined and efficient decision-making route.

Our storage framework has evolved to integrate edge relationships directly into node structures, eschewing separate edge storage. This integration is achieved through parent-child node IDs embedded in the nodes, which define the relationships. Furthermore, each node's type is explicitly identified, enhancing clarity and accessibility.

\begin{figure}
	\centering
	\includegraphics[scale=0.265]{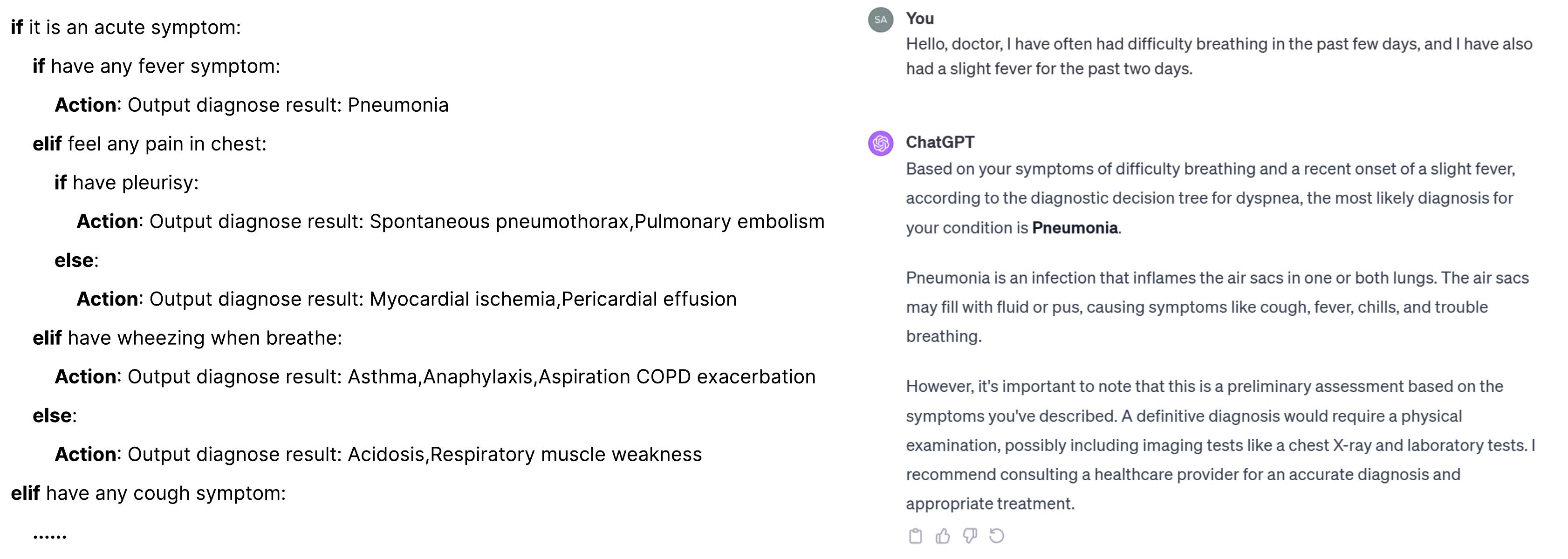}
	\caption{On the left is the clinical guidance tree using plain text structure represented by LLM-interactive IEET. And on the right is an example of feeding this IEET representation directly to ChatGPT to learn and use to diagnose a disease for a patient}
	\label{fig:figiee}
\end{figure}

\paragraph{LLM-interactive IEET.} Recognizing the difficulty of integrating our storage structure with Large Language Models (LLMs) due to potential information excess and misconceptions, we engineer an algorithm to reformat the clinical guidance tree into \textbf{LLM-interactive} \textbf{I}f-\textbf{E}lif-\textbf{E}lse \textbf{T}ree(LLM-interactive IEET) structure, depicted in Figure \ref{fig:figiee}. This format effectively presents node contents and their relationships, resonating with human cognitive patterns in clinical analysis. This streamlined and logical structure is notably more conducive to LLM comprehension and utilization than traditional structured models.

IEET structure is different from tree structure, only need to define the node content and use hierarchical order to represent the relationship between the nodes. The IEET structure, distinct from traditional tree structures, defines nodes directly without separate value attributes. And hierarchical order represents node relationships. It uses semi-structured statements of If and Elif to represent conditional nodes in CGT, and Action statements to represent action nodes. This approach emphasizes the content and interconnections of the nodes, enabling mutual conversion with standard tree structures for broader applicability.

\paragraph{Applications of LLM-interactive IEET.}The IEET structure, representing clinical guidance trees, is well-suited for interpretation by large prediction models. This allows for its direct inclusion in prompts fed into large language models, enabling these models to understand and react according to the IEET's decision process, as illustrated in Figure \ref{fig:figiee}. Furthermore, IEET structures can be converted into LLM-Executable CGT viewer formats, empowering large language models to make inference decisions based on these structures.

\begin{figure}
	\centering
	\includegraphics[scale=0.26]{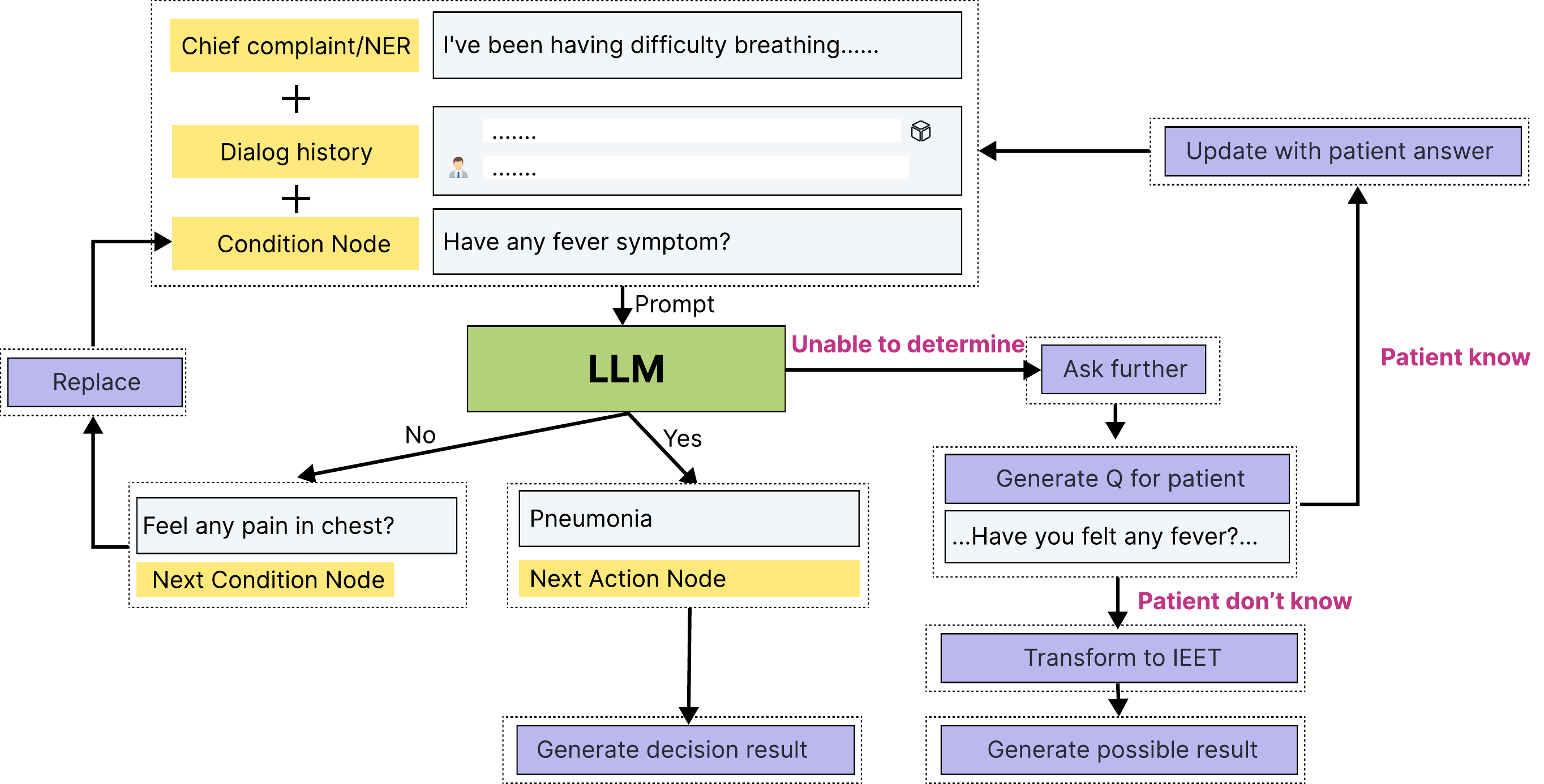}
	\caption{CDM engine: The process of inference on the CGT proposed in this paper using LLM, which can generate different responses according to different situations when making judgments on each node in the decision tree.}
	\label{fig:figreason}
\end{figure}

\subsection{LLM-executable CGT Inference Engine}
Innovatively, we proposed an inference engine based on LLM-executable CGT, named CDM engine (\textbf{C}linical \textbf{D}ecision \textbf{M}aking \textbf{Engine}). It makes use of the LLM's powerful analytical capabilities, which combine with patient information to conduct step-by-step reasoning on the decision tree, and take different measures according to different reasoning results. And not only can the engine make multi-step decisions, it can also generate follow-up questions to have multiple rounds of conversations with patients to get more patient information. Ultimately make accurate decisions for patients.

As illustrated in Figure \ref{fig:figreason}, our CDM engine involves integrating the patient's chief complaint and patient-LLM dialog history with the CGT's condition-node to formulate prompts for the large language model. This model is tasked with determining whether the complaint meets the condition-node criteria, offering one of three verdicts: 'Yes', 'No', or 'Unable to determine'. The 'Unable to determine' response is issued in cases where the model is unable to conclusively judge the compliance of the patient’s complaint with the condition-node stipulations.

Based on the judgment outcomes from the Large Language Model (LLM), the ensuing decision falls into one of three categories: Condition Node, Action Node, Ask Further. Distinct actions for different inference result are taken as follows:

\begin{itemize}
	\item \textbf{Next condition Node}. Replace the contents of the condition node and redo the reasoning in Figure \ref{fig:figreason}
	\item \textbf{Next action Node}. Reaching this node confirms a disease diagnosis using the Clinical Guidance Tree (CGT). Subsequently, the LLM is capable of producing a response informed by the node's content.
	\item \textbf{Ask Further}.Following the initial decision, we employ a Language Model (LLM) to generate a question and ask the patient about symptoms specified in the condition-node. Continue this process, facilitating a series of interactions between the LLM and the patient. This iterative dialogue ultimately leads to the diagnosis of the patient's condition. 
\end{itemize}
 
 In an actual medical consultation, patients might be uncertain about having specific symptoms, leading them to respond with "don't know" to doctors' inquiries. The CDM-engine discussed in this article addresses this by querying patients about relevant symptoms. If a patient is aware of a symptom and responds accordingly, the engine integrate the LLM-question and patient's responses into the patient-LLM dialog history. Revisit the process outlined in Figure \ref{fig:figreason} to evaluate the information. However, if a patient responds with 'I don't know', the CDM-engine adapts by transforming the current node's subtree in the CGT into the LLM-interactive IEET format introduced in this study. This approach enables the Language Model (LLM) to generate potential diagnoses by considering various possible conditions the patient might be experiencing

\section{Data Collection}
\label{sec:headings}
\begin{figure}
	\centering
	\includegraphics[scale=0.11]{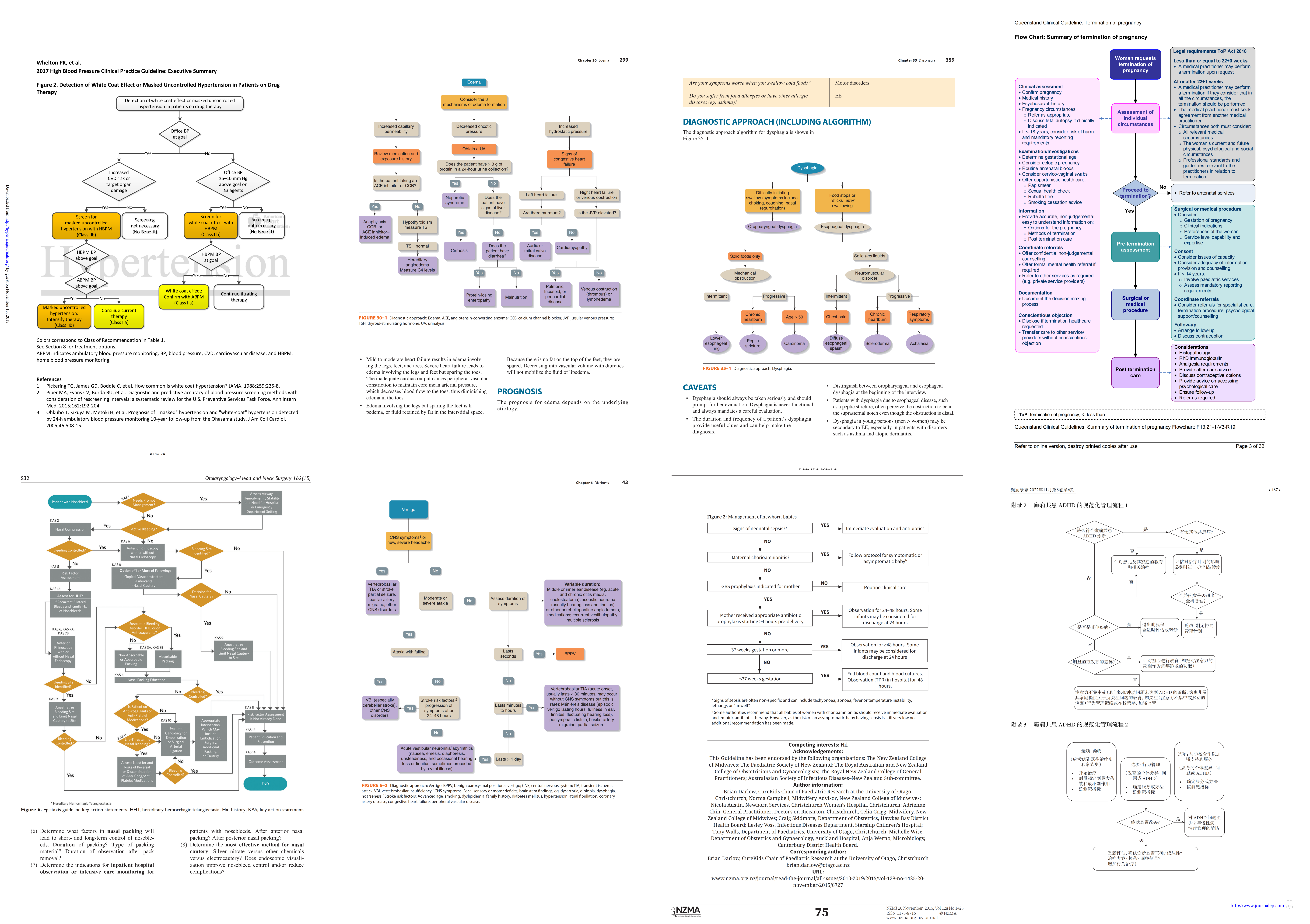}
	\caption{Include a page in the guide that contains a medical flowchart.}
	\label{fig:figflowchartpages}
\end{figure}

Our \textbf{me}dical \textbf{d}iagnostic \textbf{d}ecision-\textbf{m}aking dataset (MedDM) consists of LLM-executable CGT proposed in this paper. And clinical guidance trees were collected from numerous medical books, treatment guidelines, and other medical literature containing flowcharts. As show in figure \ref{fig:figflowchartpages}, these flowcharts clearly depict the decision-making process of doctors during diagnosis and treatment, encapsulating important knowledge of differential diagnosis, clinical decision-making, and pathological diagnosis. As show in figure \ref{fig:fig1}, this section describes the three steps to identify a flowchart: (1) collect medical books and filter out flowcharts; (2) Identify the basic shapes in the flowchart, connecting lines and words, and restructure them into a flowchart; (3) manual verification of flow chart recognition results. 

\subsection{ Medical Literature Collection and Flowchart Selection}
The knowledge in the published medical literature is often recognized in the industry or the summary of the work of some medical practitioners, which has high credibility and authority. Therefore, we collected about 5,000 published authoritative medical textbooks, treatment guidelines, and expert consensus documents to get flowcharts, covering 12 departments including emergency medicine, internal medicine, surgery, dentistry, and various diseases like gastric cancer, pneumonia, and diabetes.

\paragraph{Flowchart acquisition.} To efficiently and automatically capture all the flowcharts in medical literature, we used PyMuPDF \citet{meuschke2023benchmark} tool to load each page of the book and convert them to images. We then employed layout analysis algorithm \citet{binmakhashen2019document} of PaddleOCR to extract figures from each page other than the main text, titles, and tables. By doing so, we managed to obtain about 100,000 images. In rapid sequence, we pre-identified image using a trained Faster R-CNN model, filtering out non-flowchart images based on the criterion of having at least 8 basic flowchart shapes, eventually yielding about 2,200 flowcharts.

\paragraph{Selecting flowcharts.} To ensure the flowcharts contained sufficient medical knowledge and were clear and easy to understand, a team of five graduate students reviewed each flowchart. We set selection criteria: (1) The flowchart should begin with the disease or symptom and end with the diagnosed disease or treatment plan; (2) The decision-making process in the flowchart should be as detailed as possible and should be clear and easy to understand; (3) A flowchart cannot contain graphics other than Process, Decision, Start/End, Scan, and Arrow. After this screening, we obtained around 1202 high-quality flowcharts.

\subsection{Flowchart Recognition Pipeline}
\begin{figure}
	\centering
	\includegraphics[scale=0.6]{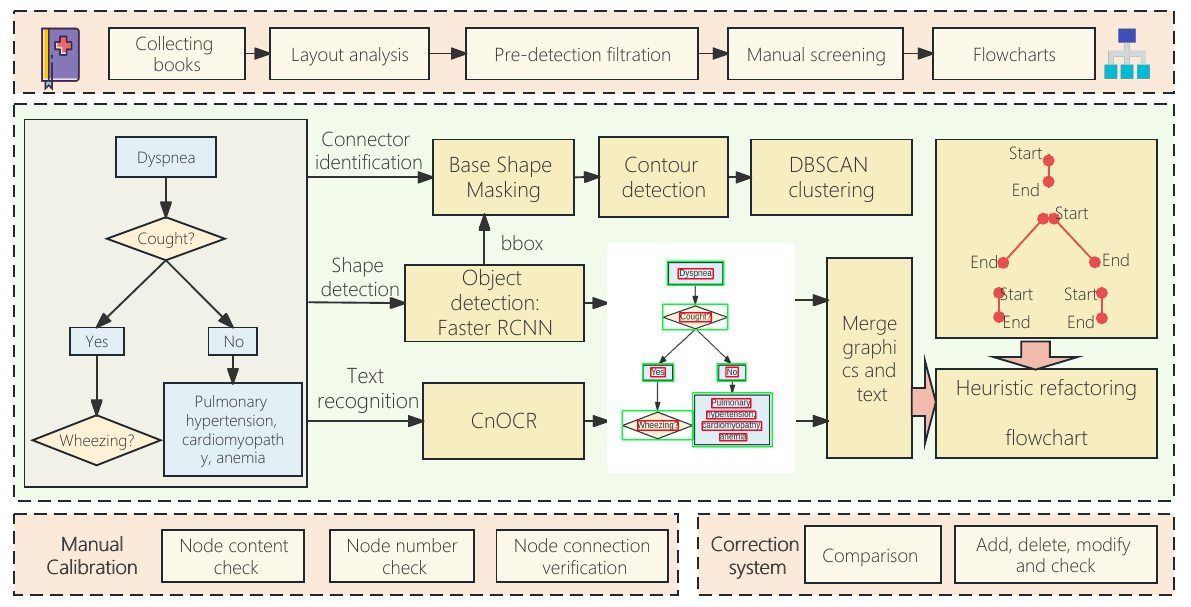}
	\caption{The overall framework for screening and identifying flowcharts from medical books.}
	\label{fig:fig1}
\end{figure}

As shown in Figure \ref{fig:fig1}, the recognition of flowcharts adopted a pipeline mode where image and connection line detection are separated. It mainly consists of four steps: 1) Shape detection, identifying basic shapes in flowcharts like Process, Decision, Start/End, Scan, and Arrow \citet{schafer2019arrow}. and detecting their position, bounding box, and classification. 2) Connection line recognition, identifying starting and ending nodes of each connecting line, which is critical for correctly representing the relationships between nodes in the flowchart. 3) Text recognition, identifying text information in the flowchart; 4) Node integration, merging each shape with its corresponding text into a node, and analyzing the connections between nodes to reconstruct a complete flowchart.

We trained a Faster R-CNN model to detect basic shapes in the flowchart. Connection line recognition used OpenCV's contour detection to obtain contour points, and DBSCAN clustering algorithm to identify starting and ending points of connection lines. Text recognition used the CnOCR model for text in basic shapes, finally reconstructing and packaging them as nodes and edges, saving the relationships between nodes and edges.

\paragraph{SHAPE DETECT.} We employed Faster R-CNN as the base shapes detection model. A dataset, called “Flowchart BS”, was created to train the model. The dataset contain 5 different classes which was generated by a shape generator. The generator draws an image with 9-16 different sizes and types of base shapes, one of which is Process, Decision, Start/End, Scan, and Arrow. To enhance detection capabilities in complex shapes, we incorporated the dataset proposed in FR-DETR \citet{sun2022fr}, excluding annotations of shapes other than the five basic types. The dataset in total comprised 2000 training images and 400 validation images. And the mAP of Faster R-CNN model achieved 97\% in basic shape detection. 


\paragraph{Connector Recognition.} To correctly identify the connections between nodes in flowcharts, we adopted a connection line recognition mode based on masking and contour analysis. As show in figure \ref{fig:fig1}, after detecting the basic shapes of the flowchart, we used the detected bounding boxes to mask basic shapes in the original image, leaving only the parts of the image with connection lines. Then, using  contour detection algorithm of OpenCV and convex hull analysis, we obtained the contour points of each connection line. Next, we used the DBSCAN clustering algorithm to aggregate the endpoints of the connection lines, determining their start and end points, thus identifying the relationships between nodes in the flowchart.

\paragraph{Text Recognition.}The medical literature involved includes two widely used languages, Chinese and English. We used the CnOCR tool for text recognition in flowcharts. The returned text information is denoted as t, t=(coordT, text), where coordT is the coordinate for locating text, and text is the recognized text content. All recognized text information in a single flowchart is denoted as T, T=(t1, t2, …, tk).

\paragraph{Reconstruct Flowchart.} After completing the recognition of shapes, connection lines, and text, we used a heuristic search \citet{segovia2021generalized} algorithm to merge the recognized images, connection lines, and text to reconstruct a complete flowchart. The steps of heuristic search are: (1) Starting from the smallest y-axis, search all texts for bounding boxes contained within the current shape's bounding box, encapsulate them into the current node, then search for the nearest starting point of the connection line. (2) Move to the endpoint of the found connection line, find the nearest shape, and construct the connection relationship between the two nodes. (3) Repeat above step until all nodes are encapsulated.

\subsection{Flowchart Calibration}
It is very important to ensure the accuracy of flowchart recognition results. Therefore, we conduct cross-validation manually. 

\paragraph{Calibration guidance.} The purpose of verification guidance is to direct personnel in verifying the recognized flowcharts, including checking the content of nodes and connection relationships. The verification criteria are: (1) The text content in each node must be accurately recognized, without typos, missing, or superfluous content. (2) The number of recognized nodes must be accurate, with no excess or missing nodes. (3) The connection relationships of the edges must be accurate, including correctly identifying one-to-many, many-to-one, and many-to-many connections.

\paragraph{Calibration system.} To more efficiently verify the recognized flowcharts, we developed a flowchart verification system. It displays all recognized flowcharts, allowing users to view the original images and the results after recognition. It supports re-identifying flowcharts by adjusting parameters and adding, deleting, modifying, and querying operations on nodes and edges of the reproduced flowcharts are allowed.

\paragraph{Formal calibration.} On account of the calibration has low requirements for calibration staff, a verification team composed of 8 graduate students was organized. Divide all flowchart into 8 parts and assigned to each person for calibration. After the calibration is completed, the two individuals exchange their respective parts to conduct the second round of verification. After two rounds of verification, the identification results can be guaranteed to be accurate.

\subsection{Flowchart transformation}
In the process of converting the identified flowchart into a CGT, there are difficulties in using the flowchart directly as a decision for doctors or systems. This is due to the flowchart having instances where the label on the connecting line is used as a node representation, the process of loop judgment, and situations where multiple nodes lead to the same node. Therefore, in our conversion process, we have specifically addressed these three aspects. 
\paragraph{Reconstructed label.}  There are a large number of processes that represent the "yes" and "no" of the conditional judgment result as a single node, rather than as a label on the connector. 

\paragraph{Elimination cycle.}  There is a loop structure in the flowchart to repeat the judgment, but the loop structure may cause the multi-tree to fail into a dead loop. We represent the flowchart using a preused multitree structure, and then use the DFS algorithm to traverse each path $P=\{N_1, N_2, \ldots, N_k\}$. When node $N_{k+1}$ traversed exists in the path $P$, it means that a loop has been entered. At this time, the relationship between $N_k$ and $N_{k+1}$ is removed, a new node $N_{n+1}$ is added, and $N_{n+1}$ is regarded as the child node of $N_k$. 

\paragraph{Modification relation.}  Commonly, there is no common child node in the tree. However, there is a situation in the flowchart where the results of multiple nodes all lead to the same node. There are node sets $C=\{N_{i+1}, N_{i+2}, \ldots, N_{i+m}\}$, the next node is also node $N_j$. We replicate node $N_j$ as children of every node in $C$. 

\subsection{Validation}
To ensure the accuracy of the transformation results, a manual verification team was also established to verify the transformation results. We organized seven graduate students to use the validation system in order to check whether there are still problems in the generated clinical guidance tree with conditional judgment results expressed as nodes, cyclic judgment structures, and overlapping paths. Once such problems are identified, they are corrected through the validation system.

\subsection{Data statistics}
\begin{table}
	\caption{The distribution of the two CGTs of differential diagnosis and treatment recommendations in each department}
	\centering
	\begin{tabular}{lllll}
		\toprule
				& Differential Diagnosis     & \%     & Treatment suggestion & \% \\
		\midrule
		Department of Internal medicine &	167&	37.6 &	36 &	4.7 \\
		Department of surgery &	59&	13.3 &	6 &	0.8 \\
		Department of pediatrics &	5&	1.1 &	52 &	6.8 \\
		Department of Obstetrics and gynecology &	7&	1.5 &	131 &	17.2 \\
		Department of Emergency &	72&	16.3 &	12 &	1.6 \\
		Department of psychiatry &	2&	0.5 &	18 &	2.4 \\
		Department of Anesthesiology &	28&	6.3 &	221 &	29.1 \\
		Department of Dermatology &	2&	0.5 &	1 &	0.1 \\
		Department of Five Senses &	79&	17.8 &	119 &	15.7 \\
		Department of Oncology &	10&	2.3 &	110 &	14.5 \\
		Department of Infectious Diseases &	7&	1.6 &	30 &	3.9 \\
		Preventive Care Division &	5&	1.1 &	23 &	3.0 \\
		Total &	443 &  &  759 &  \\	
		\bottomrule
	\end{tabular}
	\label{tab:table}
\end{table}

Table~\ref{tab:table} provides statistics of the clinical guidance tree we extract from flowcharts in books. There are totally 1202 decision trees from different departments. Based on the primary purpose of CGT, these were categorized as differential diagnosis or treatment recommendation. The differential diagnosis trees, which are totally 443, are designed to aid in pinpointing the patient's disease using a systematic approach. And treatment recommendation trees, which is totally 759, focus on deciding treatment strategies for patients.

Table~\ref{tab:table} presents the allocation of two distinct types of clinical guidance trees across various hospital departments, encompassing a total of 12 different departments. Notably, within the realm of internal medicine, differential diagnosis CGTs are predominant, amounting to 167 in total. Conversely, in the department of anesthesiology, the focus shifts to treatment recommendation CGTs, where they are most prevalent with a count of 221.

\section{Dialog framework based on LLM-executable CGT}
Based on the LLM-executable CGT constructed in the above section, owing to its encompassment of rich differential diagnosis and treatment plan knowledge, it serves as an external knowledge base aiding the LLM in emulating the logic of medical inquiries in a real manner during multi-turn dialogues with patients, thereby generating responses that are more informative and credible. The decision-retrieval based generation framework comprises two components: decision knowledge retrieval and LLM response generation.

\begin{figure}
	\centering
	\includegraphics[scale=0.2]{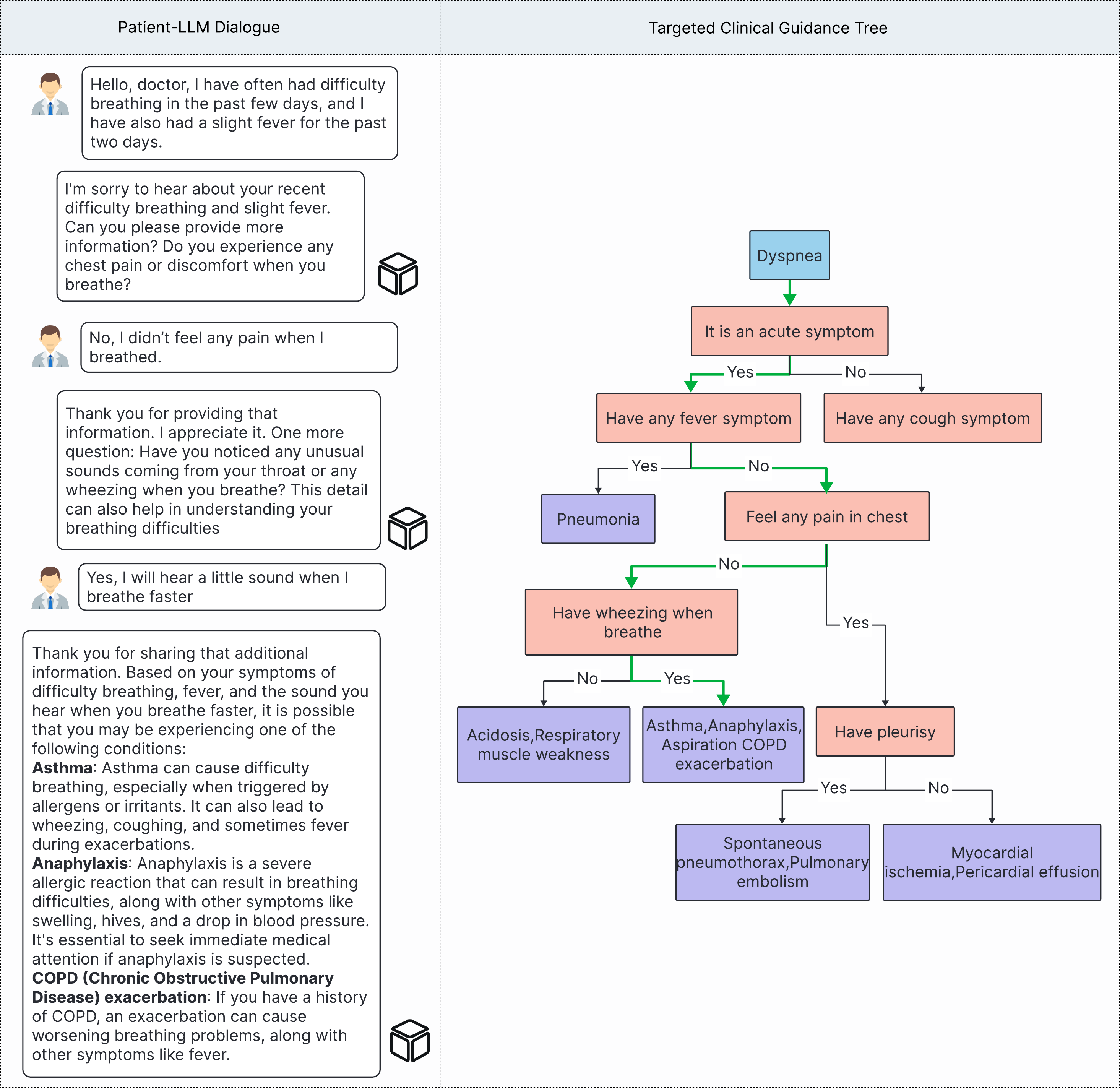}
	\caption{An example of multiple rounds of conversation between a patient and an LLM, where the LLM talks to the patient following the decision path of the CGT}
	\label{fig:figpllm}
\end{figure}

\subsection{Target CGT Retrieval}
Throughout the multi-turn interaction between the LLM and the patient, the LLM gradually collects information regarding the patient's current health status. Based on this information, the retrieval module returns diagnosis or treatment information which is relevant to the currently known information from the knowledge base.

Specifically, we vectorize the CGT for ease of computing the relevance between patient information and decision knowledge, storing it in a vector database denoted as T.

\[ T=(t_1, t_2, \ldots, t_n), t_i \in \mathbb{R}^D (1 \leq i \leq n) \]

A single query encompasses the currently known patient information, composed of historical dialogue records \( D \). However, due to the non-specialized oral expression of patients and the misalignment between medical terms in the CGT, we employ ChatGPT to rewrite the original dialogue into a more professionally articulated form, denoted as \( D' \), thereby reducing loss in the patient information description. Subsequently, we vectorize the augmented historical dialogue records and calculate the cosine similarity with \( T \) to retrieve the CGT relevant to the patient's current condition, denoted as \( t' \).

\[ t' = \max(F(q, t_i)) (1 \leq i \leq n) \]

\[ F(x, y) = \frac{x \cdot y}{|x| |y|} \]

\subsection{CGT-Retrieval Based Multi-round dialog generation}

Based on the retrieved CGT \( t' \), We follow the method described in Section 3.3 and take the LLM-executable plain text representation of each node from root as an augmentation to the external knowledge repository, denoted as \( k \). As part of the prompt, \( k \) is concatenated with the rewritten historical dialogue \( D' \), forming the complete input for the current dialogue turn. The LLM, guided by the retrieved CGT, engages in reasoning to generate the response for the current turn. The content of the response may involve further inquiry into symptoms, disclosure of disease diagnosis results, or recommendation of treatment plans, contingent upon the information provided by the patient and the CGT retrieval results.

In Figure \ref{fig:figpllm}, we present an illustration of a dialogue between a large language model and a patient. Following several judgment processes outlined in Section 3.3, the outcome of the assessment at the 'Feel any pain in chest' node is reported as 'Unable to determine.' Subsequently, this node's content is employed as a prompt for the LLM. These sophisticated models provide responses to inquiries. In the final step, the patient's responses are integrated into the primary complaint, this iterative process continues, and the patient's illness is ultimately diagnosed through multiple turns of conversations.

\section{Conclusion and Future Work}
We construct a \textbf{me}dical \textbf{d}iagnostic \textbf{d}ecision-\textbf{m}aking dataset (MedDM) extracted from flowchart in clinical guidance books. Besides, we also propose a extraction framework which is consists of two steps: flowchart recognition and decision tree construction. Using this proposed framework, we extract 1202 clinical flowcharts from over 5000 medical textbooks, clinical guidelines, consensus documents, and other relevant literature, thereby constructing the MedDM. In addition, we have defined LLM-executable clinical guidance tree, a new representation of decision trees that can be used directly for large model inference. In addition, the LLM-executable CGT reasoning framework is constructed, which can guide LLM to conduct multiple rounds of dialogue with patients according to the guideline decision-making ideas based on the proposed CGT, and finally diagnose patients' diseases. In the future, we used MedDM to generate multi-round conversation datasets to build benchmarking experiments.

\section{Limitations}
While the constructed clinical clinical guidance tree is derived from publicly available authoritative medical literature, the decision processes involved in the trees may still contain errors and outdated medical knowledge. Moreover, there is a need for further collection of medical literature for a broader coverage of medical conditions.

\bibliographystyle{unsrtnat}
\bibliography{references}

\end{document}